%% file: main.tex
\documentclass[letterpaper, 10 pt, conference]{ieeeconf}  % Comment this line out if you need a4paper

\IEEEoverridecommandlockouts                              % This command is only needed if 
\usepackage{graphics} % for pdf, bitmapped graphics files
\usepackage{amsmath,amssymb,amsfonts} % assumes amsmath package installed
\usepackage{bm}
\usepackage{subcaption}	 
\usepackage{graphicx}
\usepackage{multirow}
\usepackage{array}
\usepackage{url}
\usepackage{floatrow}
\usepackage{hyperref}
\usepackage{xcolor}
\usepackage[colorinlistoftodos]{todonotes}
\usepackage[backend=bibtex,natbib=true,style=numeric-comp,sorting=none,giveninits=true,maxbibnames=99,url=false,doi=false]{biblatex}
\title{\LARGE \bf
Targetless Calibration of LiDAR-IMU System Based on Continuous-time Batch Estimation
}

\author{Jiajun Lv, Jinhong Xu, Kewei Hu, Yong Liu, Xingxing Zuo
\thanks{ The authors are with the Institute of Cyber-Systems and Control, Zhejiang University, Hangzhou, China. (Yong Liu is the corresponding author, email: yongliu@iipc.zju.edu.cn)
}
}

\begin{document}

\maketitle
\thispagestyle{empty}
\pagestyle{empty}

%%%%%%%%%%%%%%%%%%%%%%%%%%%%%%%%%%%%%%%%%%%%%%%%%%%%%%%%%%%%%%%%%%%%%%%%%%%%%%%%
\begin{abstract}
Sensor calibration is the fundamental block for a multi-sensor fusion system. This paper presents an accurate and repeatable LiDAR-IMU calibration method (termed LI-Calib), to calibrate the 6-DOF extrinsic transformation between the 3D LiDAR and the Inertial Measurement Unit (IMU).
Regarding the high data capture rate for LiDAR and IMU sensors, LI-Calib adopts a continuous-time trajectory formulation based on B-Spline, which is more suitable for fusing high-rate or asynchronous measurements than discrete-time based approaches.
Additionally, LI-Calib decomposes the space into cells and identifies the planar segments for data association, which renders the calibration problem well-constrained in usual scenarios without any artificial targets. 
We validate the proposed calibration approach on both simulated and real-world experiments. The results demonstrate the high accuracy and good repeatability of the proposed method in common human-made scenarios. 
To benefit the research community, we open-source our code at \url{https://github.com/APRIL-ZJU/lidar_IMU_calib}

\end{abstract}

% Main sections
\input{sections/01_Introduction.tex}

\input{sections/02_Relatedwork.tex}
\input{sections/03_Trajectory_Representation.tex}

\input{sections/04_Methoodology.tex}
\input{sections/06_Experiement.tex}

\input{sections/07_Conclusion.tex}


%%%%%%%%%%%%%%%%%%%%%%%%%%%%%%%%%%%%%%%%%%%%%%%%%%%%%%%%%%%%%%%%%%%%%%%%%%%%%%%%

{
% \newpage
\vspace{0.05cm}
\def\bibfont{\scriptsize}
%================================================
\printbibliography
%================================================
%\bibliographystyle{IEEEtran}  %plainnat   %  abbrvnat   %  unsrtnat  % IEEEtranN
%\bibliography{libraries/library}
}

\end{document}

%% file: sections/01_Introduction.tex
\section{INTRODUCTION}
Multi-sensor fusion has been an essential developing trend in the robotics field. In particular, the LiDAR sensor is widely employed due to its high accuracy, robustness, and reliability, such as for localization, semantic mapping, object tracking, detection~\cite{zhang2015visual,maturana2015voxnet,Zuo2019IROS,milioto2019rangenet++,Zuo2019RAL}. 
Despite these advantages of LiDAR sensors, the downside is from the fact that the LiDAR samples a succession of 3D-points at different times, thus the motion of the sensor carrier introduces distortion like the rolling-shutter effect.
To address this issue, the Inertial Measurement Unit (IMU), widely used for the ego-motion estimation at a high frequency, can be utilized as a complementary sensor to correct the distortion.
In general, a LiDAR-IMU system benefits from the component sensors and becomes feasible for reliable perception in various scenarios.

The accuracy of ego-motion estimation, localization, mapping, and navigation are highly dependent on the accuracy of the calibration between sensors. Manually measuring the relative translation and rotation between sensors is inaccurate and sometimes impractical.
For the LiDAR-IMU calibration problem, it should be noted that the individual points in a LiDAR scan are sampled at different time instants, and the exact position of a point in the LiDAR frame is related to the pose of LiDAR at the perceived time instant. Pose estimation whenever a point is perceived is necessary for removing the motion distortion. Discrete-time based methods, like~\cite{ceriani2015pose}, leverage interpolation between poses at key time instants, which sacrifices the accuracy especially in highly-dynamic cases.  
Gentil et al.~\cite{le20183d} employ Gaussian Process (GP) regression to interpolate IMU measurements, which partially address the issue of using “low-frequency” IMU readings to interpolate for the higher-frequency LiDAR points.
Rehder et al.~\cite{rehder2014spatio} propose to calibrate the camera-IMU-LiDAR system with two stages based on continuous-time batch estimation. The camera-IMU system is calibrated with a chessboard firstly, then the single-beam LiDAR is calibrated with respect to the camera-IMU system.

 \begin{figure}[t]
     	\centering
     	\includegraphics[width=1.0\columnwidth]{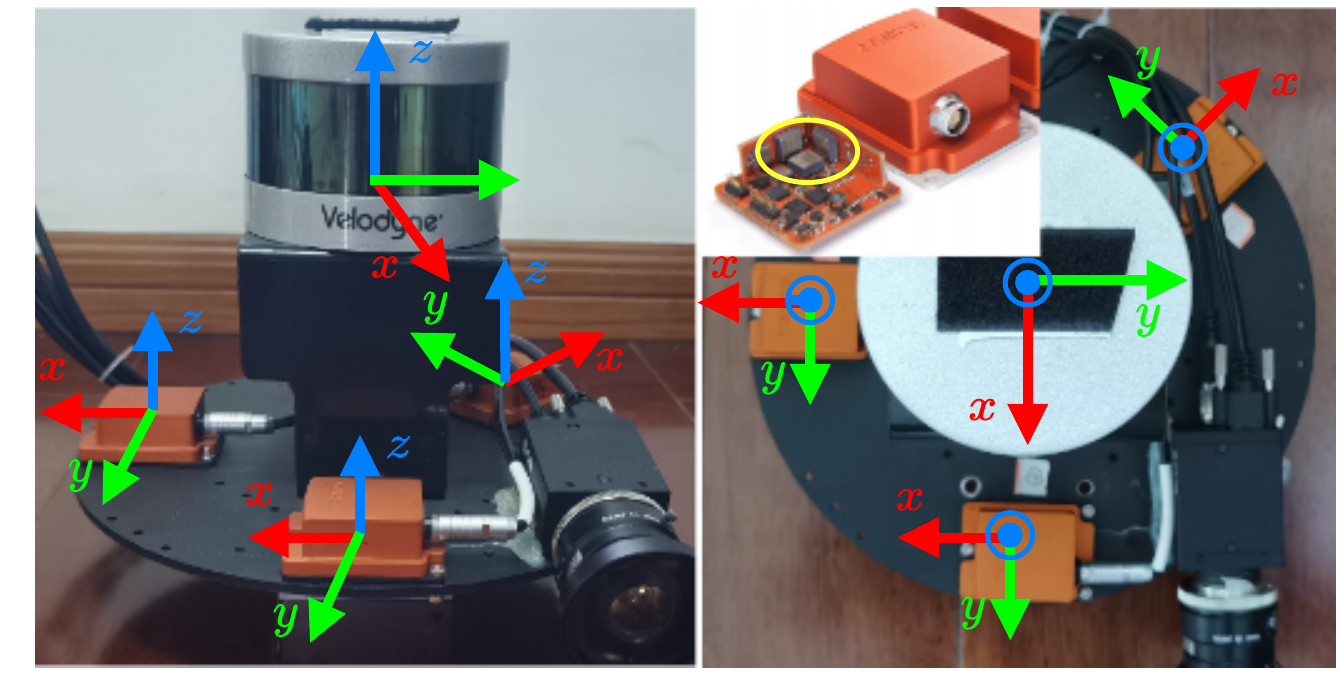}
     	\caption{The self-assembled sensors used in real-world experiments. Note that the camera is left unused in this system. The proposed methods try to calibrate the 6-DOF rigid transformation between LiDAR and IMUs.
     	}
     	\label{fig:lic_sensors}
\end{figure}

Inspired by \cite{rehder2014spatio}, we propose an accurate LiDAR-IMU calibration method (LI-Calib) in the continuous-time framework and the proposed method is an extension of our previous work~\cite{Lyu2019IROSws}.
By parameterizing the trajectory with temporal basis functions, the continuous-time based method is capable of getting exact poses at any time instants along the whole trajectory, which is ideal for the calibration problem with high-rate measurements.
Additionally, since the angular velocity and linear acceleration measurements from IMU are samples of the derivative of the temporal basis functions, the continuous-time formulation allows for a natural way to incorporate the inertial measurements.
To summarize, the main contributions of this paper are three-fold:
\begin{itemize}
	\item We propose an accurate and repeatable LiDAR-IMU calibration system based on continuous-time batch estimation without additional sensors or specially-designed targets.
	\item A novel formulation of the LiDAR-IMU calibration problem based on the continuous-time trajectory is proposed and, the residuals induced by IMU raw measurements and LiDAR point-to-surfel distances are minimized jointly in the continuous-time batch optimization, which renders the calibration problem well-constrained in general human-made scenarios.
	\item Both simulated and real-world experiments are carried out, which demonstrate the high accuracy and repeatability of the proposed method.	We further make the code open-sourced to benefit the community. To the best of our knowledge, this is the first open-sourced 3D LiDAR-IMU calibration toolbox. 
\end{itemize}

%% file: sections/02_Relatedwork.tex
\section{RELATED WORK}
Motion compensation and data association for LiDAR points are two main difficulties when calibrating the LiDAR-IMU system.
For the former problem, most existing works~\cite{ceriani2015pose,ye2019tightly} perform linear interpolation for individual LiDAR points based on the assumption that the angular velocity and the linear velocity or linear acceleration remain constant over a certain time interval.
This assumption holds for low-speed and smooth movements. When it comes to random motion (e.g., when holding the sensor with hands or when mounting the sensor on a quadrotor), it can easily be broken. 
To address this issue, Neuhaus et al.~\cite{neuhaus2018mc2slam} tightly integrate inertial data to predict the motion of the LiDAR for relaxing the linear motion assumption. Some recent works~\cite{nuchter2017improving, droeschel2018efficient, le20183d} model the trajectory with continuous-time representations that efficiently handle high-frequency distorted measurements without any assumption of the motion priors. Our proposed calibration method also employs the continuous-time formulation.
Data association for LiDAR points is the other essential problem that requires much attention. Point-to-plane and plane-to-plane data associations are utilized in ~\cite{pomerleau2015review, le20183d}. To find more reliable correspondences, Deschaud~\cite{deschaud2018imls} carries out a point-to-model strategy considering the shape information of the local neighbor points. In this paper, we adopt a special point-to-surfel data association. To be specific, the point cloud is discretized and denoted by surfels (tiny planes), and the component points are associated with its corresponding surfel.

% motion-based method
In order to calibrate a setup of rigidly-connected LiDAR and IMU, Geiger et al.~\cite{geiger2013vision} propose a motion-based calibration approach that performs the extrinsic calibration by hand-eye calibration~\cite{horaud1995hand}. However, their approach expects that each sensor's trajectory could be estimated independently and accurately, which is difficult for the commercial or consumer-grade IMU. 
Gentil et al.~\cite{le20183d} adopt Gaussian Process(GP) regression to upsample IMU measurements for removing the motion distortion in LiDAR scans and formulate the calibration as a factor-graph based optimization problem, which consists of LiDAR point-to-plane factors and IMU preintegration factors.

Continuous-time batch optimization with temporal basis functions is also widely studied in calibration problems. Furgale et al.~\cite{furgale2012continuous} detail the derivation and realization for a full SLAM problem based on B-spline basis functions and evaluate the proposed framework within a Camera-IMU calibration problem, which is further extended to support both temporal and spatial calibration~\cite{furgale2013unified}. Subsequently, Rehder et al.~\cite{rehder2014spatio} adopt a similar framework for calibrating the extrinsic between a camera-IMU system and a single beam LiDAR. And Rehder et al.~\cite{rehder2016extending} further extend the previous work~\cite{furgale2012continuous} to support multiple IMUs calibration.
In this work, we propose a continuous-time batch optimization-based LiDAR-IMU calibration method, which is similar to~\cite{rehder2014spatio} in spirit. However, chessboards and auxiliary sensors are not required in the proposed approach.

%% file: sections/03_Trajectory_Representation.tex
\section{Trajectory Representation and Notation}\label{sec:continuous_time_trajectory}

\begin{figure*}[t]
	\centering
	\includegraphics[width=1.0\textwidth]{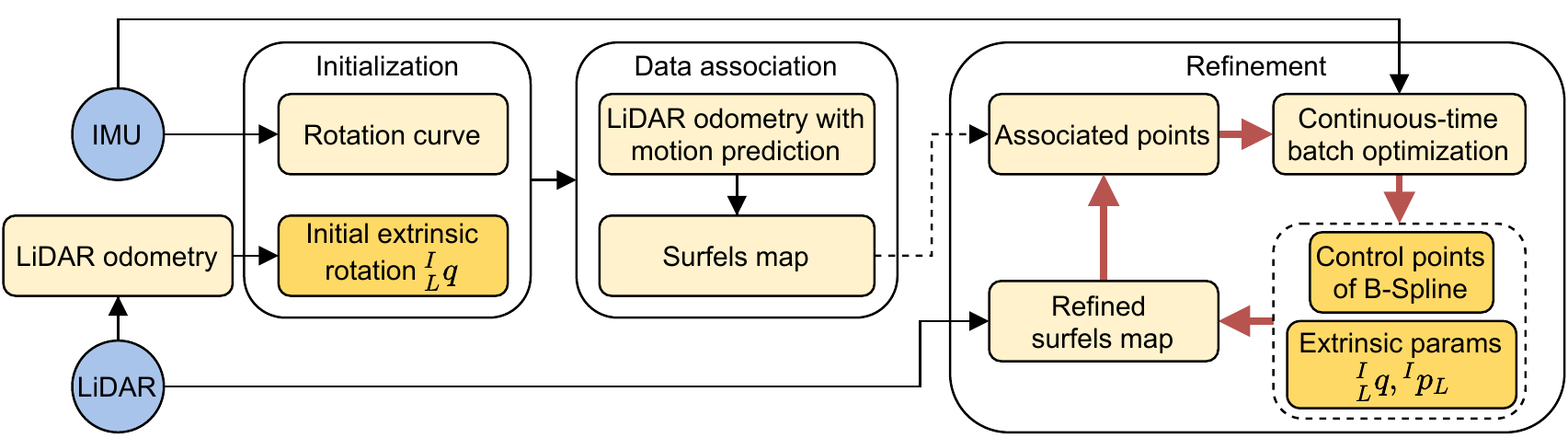}
	\caption{The pipeline of proposed LiDAR-IMU calibration method, which allows to leverage all the raw measurements from IMU and LiDAR sensor in a continuous-time batch optimization framework.}
	\label{fig:theory_flow}
\end{figure*}

The notation used in this paper can be defined as follows:
IMU frame $\{I\}$ is rigidly connected to the LiDAR frame $\{L\}$, while the map frame $\{M\}$ is located at the first LiDAR scan $\{L_0\}$ at the beginning of calibration. 
Besides, ${}^I_L\bm{q}$, ${}^I\bm{p}_L$ denote the extrinsic rotation and translation from the LiDAR frame to the IMU frame. With a little bit abuse of notation, the matrix ${}^I_L\bm{R} \in \mathit{SO}(3)$ is also the rotation corresponding to ${}^I_L\bm{q}$.

We employ B-Spline to parameterize trajectory as it provides closed-form analytic derivatives, which enables to simplify the fusion of high-frequency measurements for state estimation. B-Spline also has a good property of being locally controllable, which means the update of a single control point only impacts several consecutive segments of the spline. This trait yields a sparse system with a limited number of control points.
There are several ways to represent a rigid motion trajectory by B-Spline. Some \cite{lovegrove2013spline, mueggler2018continuous} use only one spline to parameterize poses on $\mathit{SE}(3)$ and the others~\cite{ovren2019trajectory,haarbach2018survey} use a split representation of two splines in $\mathbb{R}^3$ and $\mathit{SO}(3)$, respectively. As concluded in~\cite{haarbach2018survey}, the joint form might be torque-minimal but it is hard to control the shape of the translation curve since the translation part is tightly coupled with the rotation. Furthermore, the coupled orientations and translations are difficult to solve in our calibration problem. Consequently, we choose the split representation to parameterize the trajectory. 

The Cox-de Boor recursion formula~\cite{de1978practical} defines the basis functions of B-Spline, and a convenient way to represent splines is to use the matrix formulation~\cite{qin2000general, sommer2020efficient}. When the knots of B-Spline are uniformly distributed, B-spline is defined fully by its degree.
Specifically, for a uniform B-spline of $d$ degree, the translation $\bm{p}(t)$ over time $t\in [t_i, t_{i+1})$ is controlled only by knots at $t_i$, $t_{i+1}$, $\dots$, $t_{i+d}$ with their corresponding control points $\bm{p}_i$, $\bm{p}_{i+1}$, $\dots$, $\bm{p}_{i+d}$, and the matrix formulation of uniform B-spline could be expressed as follows:
\begin{align}
\label{Eq:p_curve}
    \bm{p}(t) = \sum_{j=0}^{d} \bm{u}^{\top}\bm{M}^{(d+1)}_{(j)} \bm{p}_{i+j}
    \text{,}
\end{align}
where $\bm{u}^{\top} = \begin{bmatrix} 1 & u & \dots & u^d \end{bmatrix}$ and $u=(t-t_i)/(t_{i+1}-t_i)$; ${\bm{M}}^{(d+1)}_{(j)}$ is the j-th column of the spline $\bm{M}^{(d+1)}$ which only depends on the corresponding degree of uniform B-Spline. In this paper, cubic ($d=3$) B-Spline is employed, thus:
\begin{align}
    \bm{M}^{(4)} =  \frac{1}{6}
                    \begin{bmatrix}  1 & 4 & 1 & 0 \\
                                    -3 & 0 & 3 & 0 \\
                                     3 & -6 & 3 & 0 \\
                                    -1 & 3 & -3 & 1 \\
                    \end{bmatrix}
    \text{.}
\end{align}
Furthermore, Eq.~(\ref{Eq:p_curve}) is equal to a cumulative representation:
\begin{align}
\label{Eq:p_curve_cumu}
    \bm{p}(t) = \bm{p}_i + \sum_{j=1}^{d} \bm{u}^{\top}\Tilde{\bm{M}}^{(d+1)}_{(j)}
     \left(\bm{p}_{i+j}-\bm{p}_{i+j-1} \right)
    \text{,}
\end{align}
where the corresponding cumulative spline matrix is as follows
\begin{align}
    \Tilde{\bm{M}}^{(4)} =  \frac{1}{6}
                    \begin{bmatrix} 6 & 5 & 1 & 0 \\
                                    0 & 3 & 3 & 0 \\
                                    0 & -3 & 3 & 0 \\
                                    0 & 1 & -2 & 1 \\
                    \end{bmatrix}
    \text{.}
\end{align}

The cumulative representation of B-spline is also widely used to parameterize orientation on $\mathit{SO}(3)$~\cite{ovren2019trajectory,haarbach2018survey,sommer2020efficient}. Kim et al.~\cite{kim1995general} first propose to use unit quaternions as the control points: 
% by changing the sum to quaternion multiplication:
\begin{align}
\label{Eq:q_curve_cumu}
    \bm{q}(t) = \bm{q}_i \otimes \prod_{j=1}^{d} exp\left(\bm{u}^{\top}\Tilde{\bm{M}}^{(4)}_{(j)}log(\bm{q}_{i+j-1}^{-1}\otimes\bm{q}_{i+j})\right)
    \text{,}
\end{align}
where $\otimes$ denotes quaternion multiplication, and $\Tilde{\bm{M}}^{(d+1)}_{(j)}$ is the j-th column of the cumulative spline matrix. 
%Thus $\bm{u}^T\Tilde{\bm{M}}^{(d+1)}_{(j)}$ is actually a scalar.
$\bm{q}_i$ is an orientation control point with its corresponding knot $t_i$, and $exp\left(\cdot\right)$ is the operation that mapping Lie algebra elements to $\mathit{S}^3$, while $log\left(\cdot\right)$ is its inverse operation~\cite{kim1995general}. 

In this calibration system, we parameterize the continuous 6-DOF poses of IMU by splines formulated by Eq.(\ref{Eq:p_curve}) and Eq.(\ref{Eq:q_curve_cumu}). The derivatives of the splines with respect to time can be easily computed \cite{sommer2020efficient}, which results in linear accelerations ${}^I\bm{a}(t)$ and angular velocities ${}^I\bm{\omega}(t)$ in the local IMU reference frame:
\begin{align}
\label{eq:spline_accel}
{}^I\bm{a}(t) &=
{}_I^{I_0}\bm{R}^{\top}(t) \left( {}_I^{I_0}\Ddot{\bm{p}}(t) - {}^{I_0}\bm{g}\right) \\
\label{eq:spline_omega}
{}^I\bm{\omega}(t) &= {}_I^{I_0}\bm{R}^{\top}(t) {}_I^{I_0}\dot{\bm{R}}(t)
\text{.}
\end{align}
In the above equation, ${}_I^{I_0}\bm{R}(t)$ represents the rotation matrix corresponding to unit quaternion ${}_I^{I_0}\bm{q}(t)$.
As described in the above equations, we set the first IMU frame $\lbrace I_0\rbrace$ as the reference frame for the IMU trajectory, and ${}^{I_0}\bm{g}$ is the gravity in the $\lbrace I_0\rbrace$ frame.

%% file: sections/04_Methoodology.tex
\section{METHODOLOGY}
\label{sec:method}

The pipeline of LI-Calib is illustrated in Fig.~\ref{fig:theory_flow}. Firstly, the extrinsic rotation ${}^I_L\bm{q}$ is initialized by aligning the IMU rotations with LiDAR rotations (Sec.~\ref{sec:rotation_init}), where the rotations of LiDAR is obtained from NDT registration~\cite{magnusson2009three} based LiDAR odometry. After initialization, we are able to partially remove the motion distortion in LiDAR scans and get better LiDAR pose estimations from LiDAR odometry. LiDAR surfels map is firstly initialized with the LiDAR poses (Sec.~\ref{sec:dataassociation}), so are the point-to-surfel correspondences.
Subsequently, continuous-time based batch optimization are conducted with the LiDAR  and IMU measurements for estimating the states including extrinsics (Sec.~\ref{sec:optimization}). Finally, with the current best estimated state from optimization, we update the surfels map, point-to-plane data association and optimize the estimated states iteratively (Sec.~\ref{sec:refine}).

\subsection{Initialization of Extrinsic Rotation}
\label{sec:rotation_init}
Inspired by~\cite{yang2016monocular}, we initialize the extrinsic rotation by aligning two rotation sequences from the LiDAR and the IMU.
Given the raw angular velocity measurements $\lbrace {}^{I_0}{\bm{\omega}}_m, \cdots, {}^{I_{M}}{\bm{\omega}}_m \rbrace$ from the IMU sensor, we are able to fit the rotational B-Splines ${}_I^{I_0}\bm{q}(t)$ in the format of Eq.(\ref{Eq:q_curve_cumu}). A series of quaternion control points of the fitted B-splines can be computed by solving the following least-square problem:
\begin{align}
\label{eq:solveq_knot}
    \bm{q}_0,\cdots,\bm{q}_N  \!=\! \arg \min \sum_{k=0}^{M}
    \left\|
    {}^{I_k}{\bm{\omega}}_m - {}_I^{I_0}\bm{R}^{\top}(t_k) {}_I^{I_0}\dot{\bm{R}}(t_k)
    \right\| \text{,}
\end{align}
where ${}^{I_k}{\bm{\omega}}_m$ is the discrete-time raw gyro measurement reported by IMU at time $t_k$.
It is important to note that the ${}^{I_0}_{I}\bm{q}(t_0)$ is fixed to an identity quaternion during the optimization of least-square.
In Eq~\ref{eq:solveq_knot}, we try to fit the rotation B-Splines by the raw IMU measurements, rather than the integrated IMU measurements (relative pose predictions), which are inaccurate and always affected by drifting IMU biases and noises. 

By leveraging scan-to-map matching with NDT-based registration, we can estimate the pose of each LiDAR scan. Thus, it is easy to get  the relative rotation between two consecutive LiDAR scans, ${}^{L_{k}}_{L_{k+1}}\bm{q}$. Besides,
the relative rotation between time interval $[t_k, t_{k+1}]$ in the IMU frame can also be obtained from the fitted B-Splines as ${}_{I_{k+1}}^{I_k}\bm{q}={}_I^{I_0}\bm{q}^{-1}(t_k) \otimes{}_I^{I_0}\bm{q}(t_{k+1})$. The relative rotations at any $k$ from two sensors should satisfy the following equation:
\begin{align}
\label{Eq:RXXB}
    {}^{I_k}_{I_{k+1}}\bm{q}\otimes {}_L^I\bm{q} 
    = {}_L^I\bm{q}\otimes {}^{L_{k}}_{L_{k+1}}\bm{q}
    \text{.}
\end{align}
By stacking multiple measurements at different time, we get the following overdetermined equation
\begin{align}
    \label{Eq:qXXB}
    \begin{bmatrix}
    \vdots\\
    \alpha_k \left(\left[ {}^{I_k}_{I_{k+1}}\bm{q} \right]_L - \left[ {}^{L_{k}}_{L_{k+1}}\bm{q} \right]_R \right) \\
    \vdots
    \end{bmatrix}  {}_L^I\bm{q} = \bm{Q}_N {}_L^I\bm{q} =\bm{0}
    \text{,}
\end{align}
where $\left[\bm{q} \right]_L$ and $\left[\bm{q} \right]_R$ are the left and right quaternion-product matrices~\cite{sola2017quaternion}, respectively; $\alpha_k$ is the weight for each rotations pair, which is determined in a heuristic way for repressing outliers: 
\begin{align}
r_k = \left\|2 \left( \arccos({}^{I_k}_{I_{k+1}}{q_w})-\arccos({}^{L_{k}}_{L_{k+1}}{q_w}) \right)\right\| \\
    \alpha_k=\left\{ \begin{array}{lr}
         1, & r_k < \text{threshold} \\
         \frac{\text{threshold}}{ r_k }, & \text{otherwise}
    \end{array}\right.
\end{align}
where ${q_w}$ is the real part of a quaternion. The solution of Eq.~(\ref{Eq:qXXB}) can be found as the right unit singular vector corresponding to the smallest singular value of $\bm{Q}_N$.

For the initialization of ${}^I\bm{p}_L$, it becomes difficult. Firstly, it should be noted that the acceleration is coupled with gravity and related to the orientation. Thus aligned error in the rotational B-Splines will affect the accuracy of translation B-Splines.
Additionally, the B-Spline's quadratic derivative introduces two zero elements in the $\bm{u}$ vector of Eq.~(\ref{Eq:p_curve}) that makes the control points unsolvable. Consequently, initializing the translation curve with raw IMU measurements is unreliable, thus we ignore the initialization of extrinsic translation, which 
has little impact on the calibration result as the experiments suggest.

\subsection{Surfels Map and Data Association}
\label{sec:dataassociation}
    
    \begin{figure}[t]
    	\begin{subfigure}{1.0\columnwidth} % width of left subfigure
    	    \centering
    		\includegraphics[width=0.8\columnwidth]{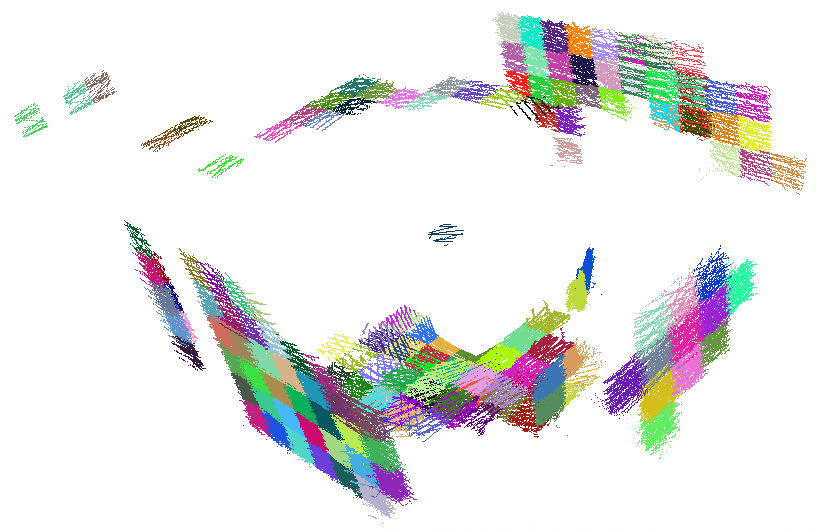}
    		\caption{The surfels map in the first iteration.} % subcaption
    		\label{fig:surfel_map1}
    	\end{subfigure}
    	%\hfill
    	\\
  	    \vspace{1em} % here you can insert horizontal or vertical space
    	\begin{subfigure}{1.0\columnwidth} % width of right subfigure
    	    \centering
    		\includegraphics[width=0.8\columnwidth]{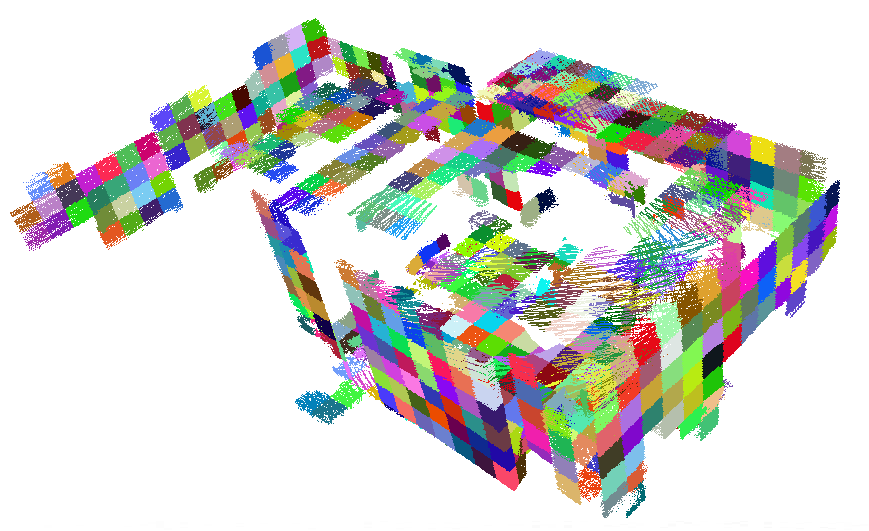}
    		\caption{The surfels map in the second iteration.} % subcaption
    		\label{fig:surfel_map2}
    	\end{subfigure}
    	\caption{Typical surfels map in an indoor scenario. The initial surfels map suffers from motion distortion with the initialized LiDAR poses. With the estimated states after just one iteration of batch optimization, the quality of refined surfels map becomes significantly better. Points in different surfels are colorized differently.} % caption for whole figure
    	\label{fig:surfel_map}
    \end{figure}

With the initialized ${}_L^I\hat{\bm{q}}$ and orientation curve, the IMU sensor can provide rotation predictions for the LiDAR scan registration and remove rotational distortion in raw scans, which improves the accuracy of LiADR odometry and the quality of LiDAR point cloud map.
Subsequently, we discretize the LiDAR point cloud map into 3D cells and calculate the first and second order moments of the points insides each cell. Surfels can be identified based on the plane-likeness coefficient~\cite{bosse2009continuous}:
\begin{equation}
    \mathcal{P}=2 \frac{\lambda_{1}-\lambda_{0}}{\lambda_{0}+\lambda_{1}+\lambda_{2}}
    \text{,}
\end{equation}
where $\lambda_{0} \leq \lambda_{1} \leq \lambda_{2}$, and they are eigenvalues of the second-order moment matrix. For the cell with planar distribution, $\mathcal{P}$ will be close to 1. Therefore, if the plane-likeness $\mathcal{P}$ is larger than a threshold, we will fit a plane $\pi$  in this cell by using the RANSAC algorithm. The plane $\pi$ is denoted by its normal vector and the distance from origin $\pi=\left[\bm{n}^{\top}, d\right]^{\top}$. 
It should be noted that both the planes and the LiDAR map are denoted in frame $L_0$.
As described above, we extract tiny planes from small cells rather than big planes in the environments, which allows us to make full use of all planar areas in the environment and provide more reliable constraints for batch optimization. 
For robustness, we reject the point-to-plane correspondences when the distance is over a certain threshold. To limit the computational cost, we also downsample the raw LiDAR scans by random sampling.  

\subsection{Continuous-time Batch Optimization}
\label{sec:optimization}
The estimated state in the system  can be summarized as:
\begin{equation}
    \bm{x} = [ {}^I_L\bm{q}^\top, {}^I\bm{p}_L^{\top}, \bm{x}_q^\top, \bm{x}_p^\top, \bm{b}_{a}^\top, \bm{b}_{a}^\top ]^{\top} 
    \text{,}
\end{equation}
where $\bm{x}_q = [\bm{q}_0^\top,\cdots,\bm{q}_N^\top]^\top$ and $\bm{x}_p = [\bm{p}_0^\top,\cdots,\bm{p}_N^\top]^\top$ denote all the control points for the rotation and translation of continuous-time trajectory described in Sec.~\ref{sec:continuous_time_trajectory}. ${}^{I_0}_{G}\bar{\bm{q}}$ is a two-dimension rotation matrix that aligns the $z$ axis of IMU reference frame ${I_0}$ to the gravity.
The bias of the accelerometer and gyroscope, $\bm{b}_{a}$ and $\bm{b}_{g}$, are assumed to be affected by white noises. 
In general, the calibration problem can be formulated as a graph-based optimization problem.
Given a sequence of associated LiDAR points $\mathcal{L}$, linear acceleration measurements $\mathcal{A}$, angular velocity measurements $\mathcal{W}$, and the assumption of independent Gaussian noise corruption on all the above measurements, the maximum likelihood estimation problem $p\left(\bm{x}|\mathcal{L},\mathcal{A}, \mathcal{W}\right)$ can be approximated by solving the following least-square problem:
\begin{align}
\label{Eq:problem}
    \hat{\bm{x}} \!=\! \arg \min \left\{
    \sum_{k \in  \mathcal{A}} \left\|\bm{r}^{k}_{{a}}\right\|_{\bm{\Sigma}_{{a}}}^{2}
    +\sum_{k \in \mathcal{W}} \left\|\bm{r}^{k}_{{\omega}}\right\|_{\bm{\Sigma}_{{\omega}}}^{2}
    +\sum_{j \in \mathcal{L}} \left\|\bm{r}^{j}_{\mathcal{L}}\right\|_{\bm{\Sigma}_{\mathcal{L}}}^{2}
    \right\}
    \text{,}
\end{align}
where $\bm{r}^{k}_{{a}}$, $\bm{r}^{k}_{{\omega}}$, $\mathbf{r}^{j}_{\mathcal{L}}$ are the residual errors associated to the accelerometer, gyroscope and LiDAR measurements, respectively. $\bm{\Sigma}_{{a}}$, $\bm{\Sigma}_{{\omega}}$, $\bm{\Sigma}_{\mathcal{L}}$ are the corresponding covariance matrices. Particularly, the IMU residuals are defined as:
\begin{align}
\label{Eq:ra}
    \bm{r}^{k}_{{a}} & = {}^{I_k}{\bm{a}}_m - {}^{I}\bm{a}(t_k) - \bm{b}_a
\end{align}
\begin{align}
\label{Eq:rw}
    \bm{r}^{k}_{{\omega}} & =  {}^{I_k}{\bm{\omega}}_m - {}^{I}{\bm{\omega}}(t_k) - \bm{b}_g
    \text{,}
\end{align}
where ${}^{I_k}{\bm{a}}_m$ and ${}^{I_k}{\bm{\omega}}_m$ are the raw IMU measurements at time $t_k$.
For a associated LiDAR point ${}^{L_{j}}\bm{p}_i \in \mathcal{L}$, captured at time $t_j$ and associated with plane $\pi_i$, the point-to-plane distance can be calculated as follows:
\begin{align}
    {}^{L_{0}}\bm{p}_i &= 
    {}_L^{I}\bm{R}^\top {}_{I_j}^{I_0}\bm{R} {}_L^{I}\bm{R}  {}^{L_{j}}\bm{p}_i + {}^{L_{0}}\bm{p}_{L_j} \notag \\ 
    {}^{L_{0}}\bm{p}_{L_j} &= {}_L^{I}\bm{R}^\top {}_{I_j}^{I_0}\bm{R} {}^I\bm{p}_L + {}_L^{I}\bm{R}^\top {}^{I_0}\bm{p}_{I_j} - {}_L^{I}\bm{R}^\top {}^I\bm{p}_L \notag
    \\
    \mathbf{r}^{j}_{\mathcal{L}} &= 
    \begin{bmatrix}
        {}^{L_{0}}\bm{p}^\top_i &    1 
    \end{bmatrix} \pi_j
    \text{.}
\end{align}
We adopt Levenberg-Marquardt method to minimize Eq.(\ref{Eq:problem}) and solve the estimated states iteratively.
\iffalse
we first transform the point to the map frame which coincides with the first scan 
\begin{align}
\label{eq:LkT0L0}
    {}^{I_{0}}\bm{x}_j &= 
    {}_I^{I_0}\bm{R}(\tau_j) \left({}_L^I\bm{R}  {}^{L_{j}}\bm{x}_j + {}^I\bm{p}_L\right) + {}^{I_0}\bm{p}_I(\tau_j) \notag \\
    {}^{L_{0}}\bm{x}_j &= {}_L^I\bm{R}^{\top}\left( {}_I^{I_0}\bm{R}^{\top}(\tau_0) \left( {}^{I_{0}}\bm{x}_j - {}^{I_0}\bm{p}_I(\tau_0) \right) - {}^I\bm{p}_L \right)
    \text{,}
\end{align}
where $\tau_0$ is the timestamp of the first LiDAR point in the first scan. Then the LiDAR residual can be formulated as the point-to-surfel distance as follows:
\begin{align}
    \mathbf{r}^{j}_{\mathcal{L}} = \pi^T_j 
    \begin{bmatrix}
        {}^{L_{0}}\bm{x}_j \\ 
        1 
    \end{bmatrix}
    \text{.}
\end{align}
We adopt the Levenberg-Marquardt method to solve this least square minimization problem attractively.
\fi

\subsection{Refinement}
\label{sec:refine}
After the continuous-time batch optimization, the estimate states including the extrinsics become more accurate. Thus, we leverage the current best estimates $\hat{\bm{x}}$ to remove the motion distortion in the raw LiDAR scans, rebuild the LiDAR surfels map, and update the point-to-surfel data associations. Note that, NDT registration based LiDAR odometry is only utilized at the very beginning (first iteration in batch optimization) for initializing the LiDAR poses and LiDAR map .
The typical LiDAR surfels maps in the first and second iteration of the batch optimization are shown in Fig.~\ref{fig:surfel_map}. The quality of the map can be improved in a significant margin after just one iteration of batch optimization. With only few iterations (around 4 iterations in our experiments), the proposed method converges and is able to give calibration results with high accuracy.

%% file: sections/06_Experiement.tex
\section{EXPERIMENTS}
%%%%%%%%%%%%%%%%%%%%%%%%% simulation figure
    \begin{figure} [t]
		\centering
		\begin{subfigure}{0.49\columnwidth} % width of left subfigure
		    \centering
			\includegraphics[height=0.8\columnwidth]{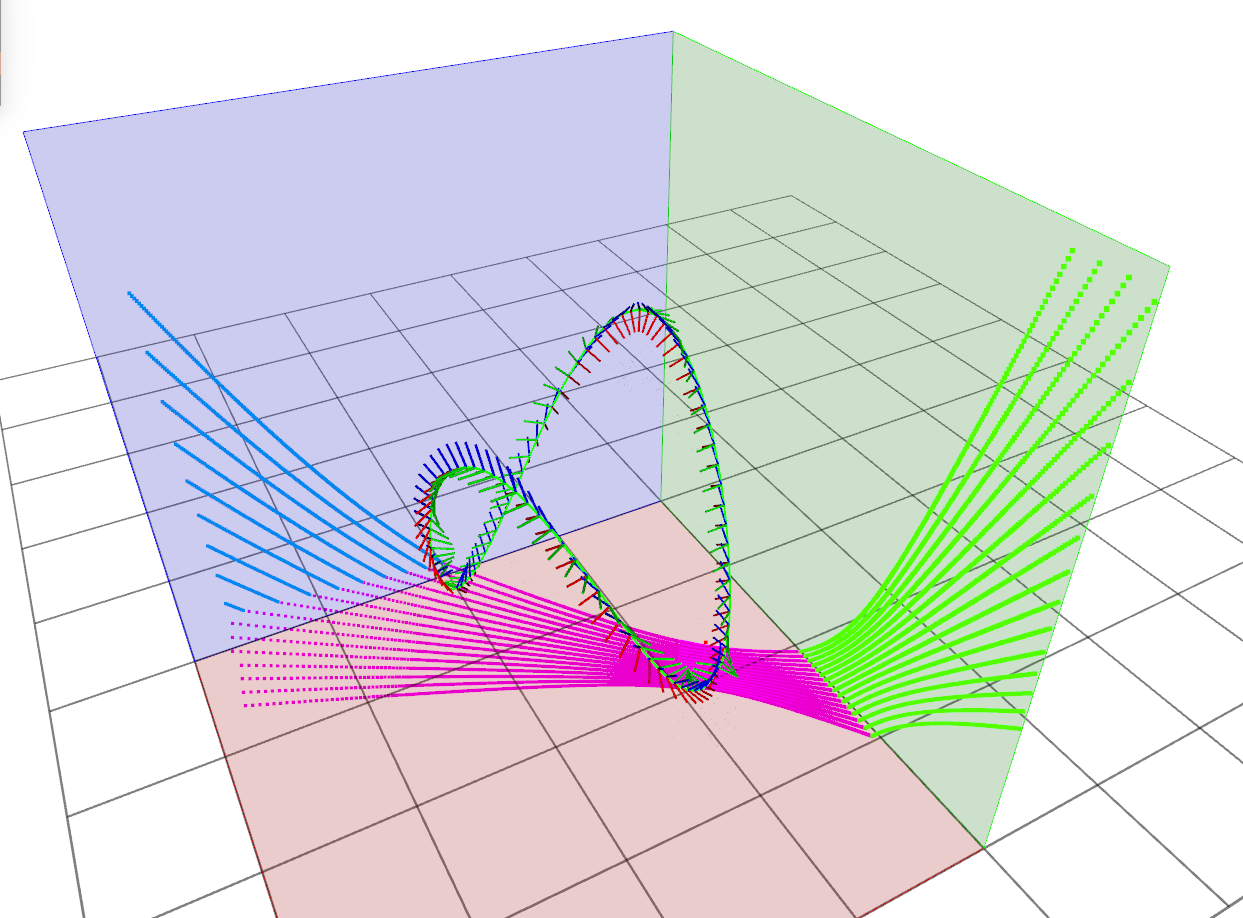}
		\end{subfigure}
		\begin{subfigure}{0.49\columnwidth} % width of right subfigure
		    \centering
			\includegraphics[height=0.8\columnwidth]{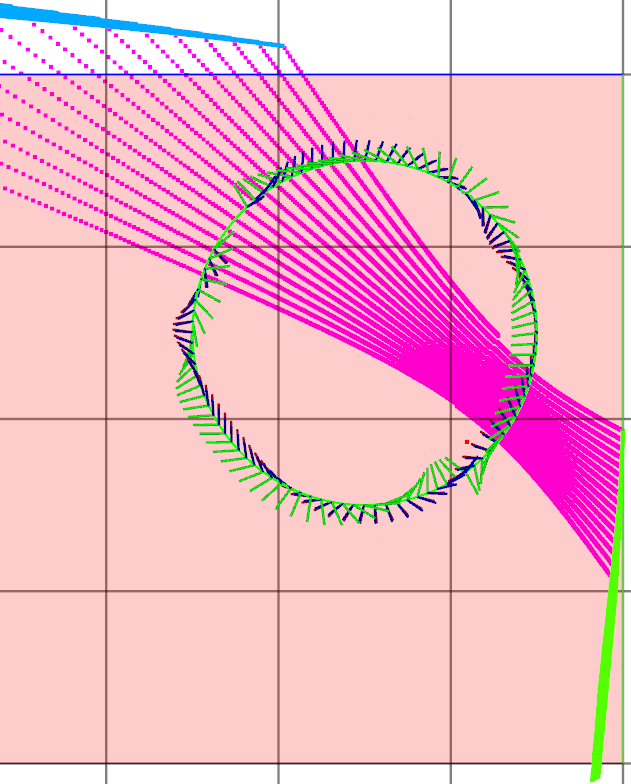}
		\end{subfigure}
		\caption{In the simulation, three planes are orthogonal to each other. The IMU trajectory is denoted by sequential coordinate markers, and the discrete colored points are the LiDAR scans with motion distortion. The size of the background grid is 2m. 
		}
		\label{fig:simulation}
	\end{figure}

%%%%%%%%%%%%%%% vicon data
	\begin{figure}[t]
		\centering
		\begin{subfigure}{0.49\columnwidth} % width of left subfigure
			\includegraphics[width=\columnwidth]{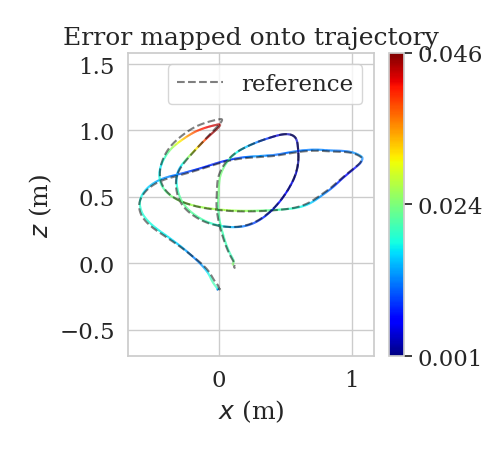}
		\end{subfigure}
		\hfill	
		\begin{subfigure}{0.49\columnwidth} % width of right subfigure
			\includegraphics[width=\columnwidth]{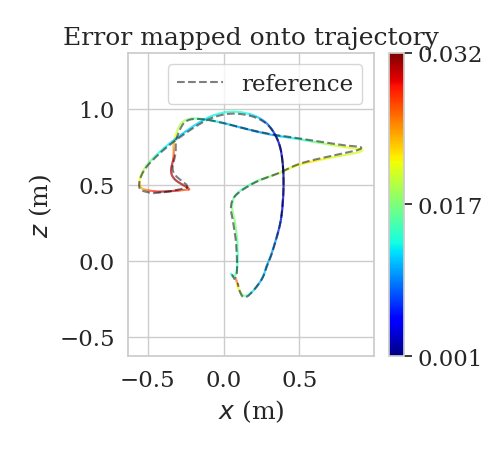}
		\end{subfigure}
		
		\begin{subfigure}{1.0\columnwidth} % width of right subfigure
			\includegraphics[width=\columnwidth]{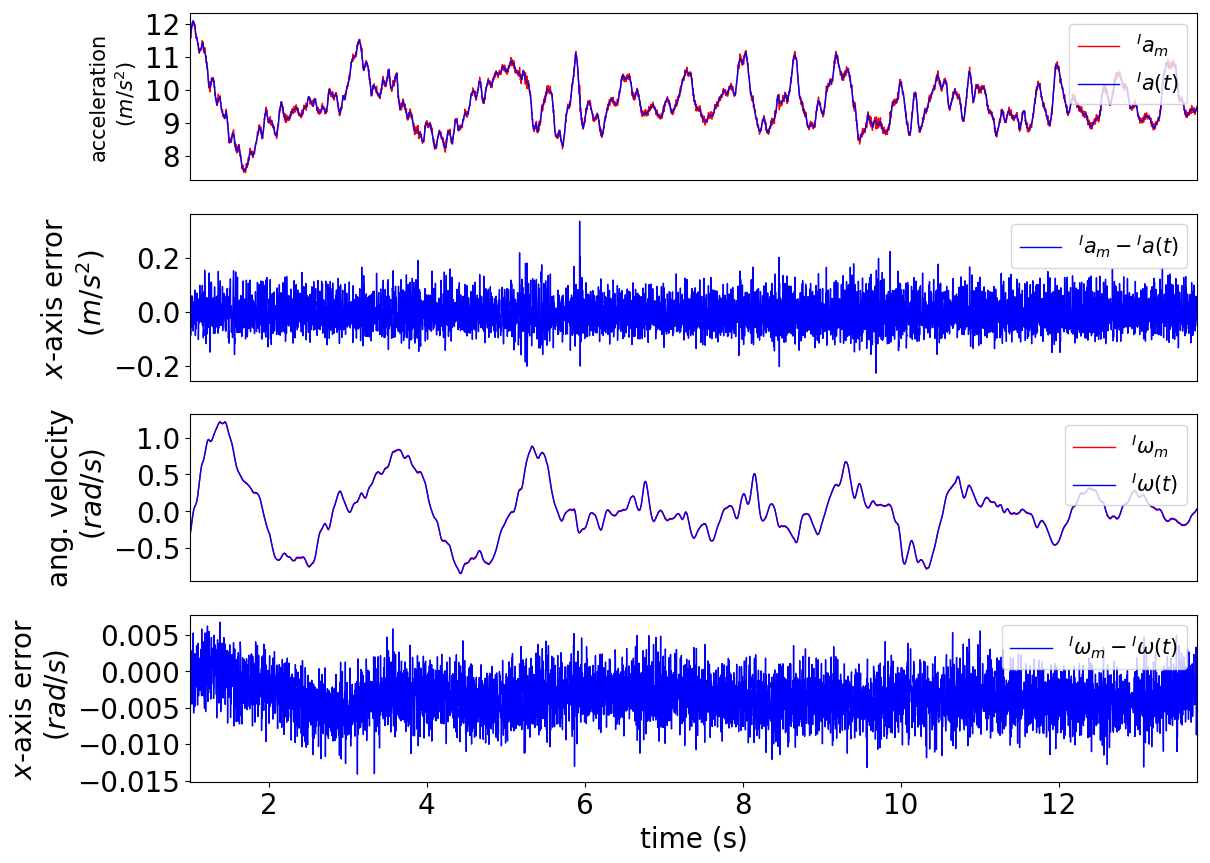}
		\end{subfigure}
		\caption{Top: Estimated continuous-time trajectories aligned with the ground-truth. The color of the trajectory indicates ATE. Bottom: The corresponding fitting results of the top-left trajectory, according to Eq.~(\ref{Eq:ra}) and Eq.~(\ref{Eq:rw}). Only the $x$-axis is plotted. The fitting errors almost have the same order of magnitude as the IMU measurement noise.} % caption for whole figure
		\label{fig:trajApe}
	\end{figure}

%%%%%%%%%%%%%%%%%%% real world experiments - the iteration process    
	\begin{figure}[t]
	\centering
	\begin{subfigure}{1.0\columnwidth} % width of left subfigure
		\includegraphics[width=\columnwidth]{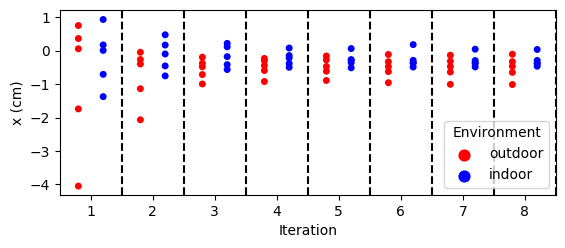}
	\end{subfigure}
	\hfill	
	\begin{subfigure}{1.0\columnwidth} % width of right subfigure
		\includegraphics[width=\columnwidth]{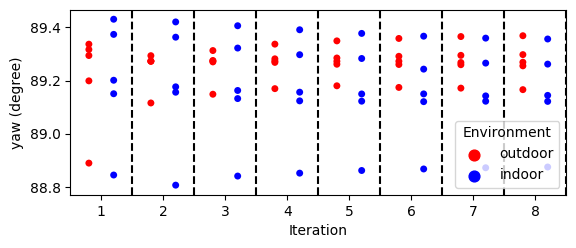}
	\end{subfigure}
	\caption{As the number of iterations increases, both translation and rotation are gradually converging, after about four iterations, the results gradually stabilized and remained unchanged.
	Comparing the difference between the indoor and outdoor calibration results, we can find that better calibration results can be obtained in indoor environments with more flat surfaces.}
	\label{fig:8iterration}
\end{figure}
%%%%%%%%%%%%%%%%%%% real world experiments - the cad visualization
    \begin{figure}[t]
    	\begin{subfigure}{1.0\columnwidth} % width of left subfigure
    	    \centering
			\includegraphics[width=\columnwidth]{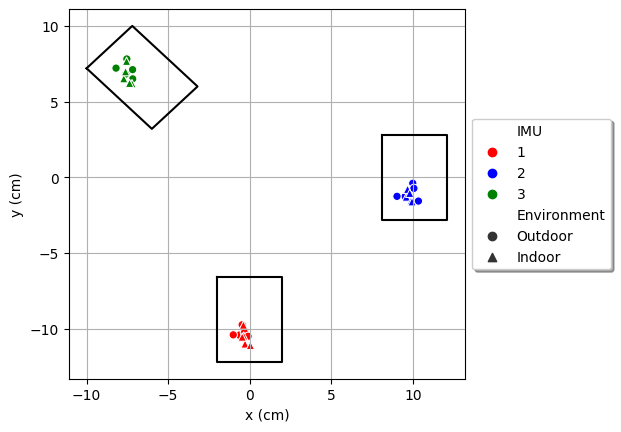}
		\end{subfigure}
		\hfill
		\begin{subfigure}{1.0\columnwidth} % width of right subfigure
		    \centering
			\includegraphics[width=\columnwidth]{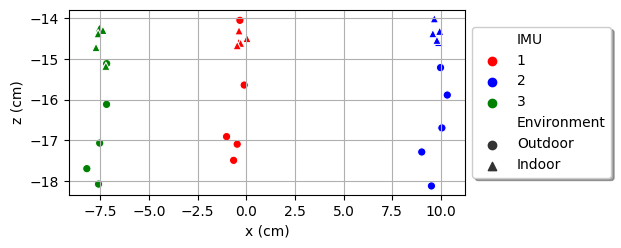}
	\end{subfigure}
		
    	\caption{The calibration results of the sensor setup shown in Fig.~\ref{fig:lic_sensors}  after the 8th iteration. Top: Top view of the calibration results. The three rectangular boxes represent the actual installation positions of the three IMUs, respectively.
    	Bottom: Front view of the calibration results. The calibration results over the indoor data are generally more reliable than the results over the outdoor data.
    	}
    	\label{fig:cad}
    \end{figure}

To validate the proposed method, we conduct extensive experiments on both simulated and real-world datasets.
We begin the evaluations with simulation studies, in which we try to analyze the calibration accuracy achieved by LI-Calib.
As for the real-world experiments, we try to calibrate the self-assembled sensors with a LiDAR VLP-16\footnote{https://velodyneLiDAR.com/vlp-16.html} and three Xsens IMUs~\footnote{https://www.xsens.com/products/mti-100-series/}, as shown in Fig.~\ref{fig:lic_sensors}. 
In the experiments, the LiDAR and IMU are synchronized by hardware~\cite{Xu2018RealTimeLD}, and we only focus on the spatial calibration between two sensors. Furthermore, in our implementation, Kontiki~\cite{kontiki} toolkit is adopt for the continuous-time batch optimization.

\subsection{Simulation}
In the simulation, the simulated sensors' characteristics are consistent with the actual sensors used in the real-world experiments. IMU measurements are reported in 400Hz. LiDAR scans are received in 10Hz with a FOV of 360${}^\circ$ horizontally, and $\pm$15${}^\circ$ vertically.  
Fig.~\ref{fig:simulation} shows the simulation environment, where three planes are orthogonal to each other, and the IMU moves in a sinusoidal trajectory. 
We conduct Monte Carlo experiments by simulating 10 different calibration sequences with the same time span of 10 Sec. Gaussian noises with the same characteristics of real-word sensors are generated and added into the synthetic measurements.
Compared to the ground truth, the reported calibration results are with translational errors of 0.0043 $\pm$ 0.0006 meters and orientational errors of 0.0224 $\pm$ 0.0026 degrees, which demonstrates the high accuracy and effectiveness of the proposed LI-Calib.

\subsection{Real-world Experiments}
%%%%%%%%%%%%%%% realWorld data
\begin{figure}[t]
    	\centering
    	\includegraphics[width=1.0\columnwidth]{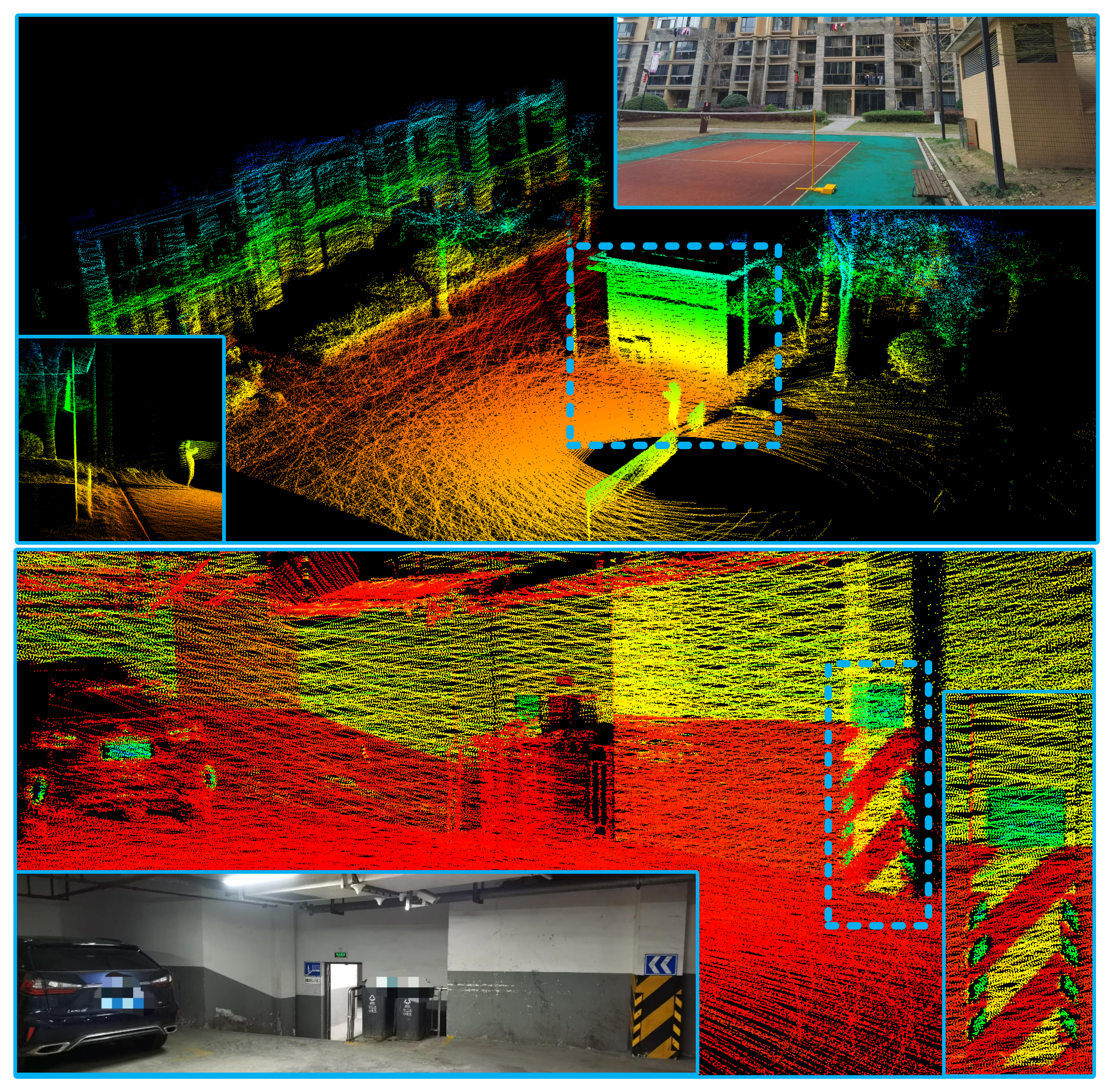}
    	\caption{The two different scenarios in real-world experiments. The point cloud maps consist of all undistorted LiDAR points by the calibration results. The top one is colored with height, and the bottom is with reflective intensity for clarity. }
    	\label{fig:realWorld}
\end{figure}
    
%%%%%%%%%%%%% Dataset and hardware
We collected data in both indoor and outdoor scenarios (see Fig.~\ref{fig:realWorld}) by the self-assembled sensors in Fig.~\ref{fig:lic_sensors} for the real-world experiments. In each scenario, five sequences with a 25-30 Sec duration are recorded with an average absolute angular velocity of about $47.39 ^\circ/s$ and an average absolute acceleration of $0.628 m/s^2$ overall sequences.
Besides, we also collect ten sequences in the Vicon room with a duration of about 15 Sec to evaluate the accuracy of estimated trajectories and the representation capability of the splines. 
%%%%%%%%%%%%% Evaluation method
Due to the absence of ground-truth extrinsic transformation between LiDAR and IMU in real-world experiments and the \textbf{lack of open-sourced LiDAR-IMU calibration algorithms}, the relative poses of three IMU inferred from CAD assembly drawings are introduced as references, and the repeatability of the calibration results is also examined over different calibration sequences. 
Furthermore, we also analyze the results computed from different numbers of iterations in the refinement step.

%%%%%%%%%%%%%%% 3 XSENS tables
\begin{table*}[ht]
	\caption{The mean and SD(standard deviation) of the calibration results for five different trails in two scenarios. 
	} %\vspace{-1em}
	\label{table:mean-std-results}
	\centering
		\resizebox{\linewidth}{!}
		{
		    \begin{tabular}{l|c|c|c|c|c|c|c}
    		    \hline
    			& &x$(m)$ & y$(m)$ &z$(m)$ &roll$({}^\circ)$ &pitch$({}^\circ)$ &yaw$({}^\circ)$\\
     			\hline
     			\multirow{3}{*}{Indoor Scene} & IMU1 &$-0.0030\pm \textbf{0.0020}$ &$-0.1058\pm0.0053$ &$-0.1453\pm0.0014$ &$180.12\pm \textbf{0.18} $ &$179.81\pm \textbf{0.28} $ &$89.15\pm \textbf{0.18}$ \\
     			\cline{2-8}
     			& IMU2 & $0.0978\pm0.0015$  & $-0.0127\pm0.0035$ & $-0.1437\pm0.0023$ & $180.05\pm0.06$ & $179.63\pm0.21$ & $88.48\pm0.11$ \\
     			\cline{2-8}
     			& IMU3 & $-0.0749\pm0.0020$ & $0.0671\pm\textbf{0.0062}$  & $-0.1456\pm \textbf{0.0039}$ & $-0.35\pm0.17$  & $0.61\pm0.20$   & $133.20\pm0.17$ \\
    			\hline
    			\multirow{3}{*}{Outdoor Scene} & IMU1 & $-0.0051\pm0.0034$ & $-0.1027\pm0.0031$  & $-0.1623\pm \textbf{0.0140}$ & $179.99\pm0.07$  & $179.82\pm0.29$   & $89.27\pm0.12$ \\
    			\cline{2-8}
    			& IMU2 & $0.0978\pm \textbf{0.0052}$ & $-0.0105\pm0.0047$  & $-0.1664\pm0.0114$ & $179.91\pm0.15$  & $179.64\pm \textbf{0.36}$   & $88.65\pm0.10$ \\
    			\cline{2-8}
    			& IMU3 & $-0.0753\pm0.0042$ & $0.0709\pm \textbf{0.0049}$  & $-0.1681\pm0.0120$ & $-0.24\pm \textbf{0.29}$  & $0.56\pm0.26$   & $133.35\pm \textbf{0.13}$ \\
    			\hline
    		\end{tabular}
		}
\end{table*}

 \begin{table}[ht]
	\caption{
	The relative rotation and translation of IMU2 (or IMU3) w.r.t. IMU1 compared with CAD reference over the indoor datasets.
	}
	\label{table:cad-results}
	\centering
		\resizebox{\linewidth}{!}
		{
		    \begin{tabular}{l|c|c|c}
		        \hline
    			&Estimated$(m)$ &  CAD Reference$(m)$ &Difference$(m)$ \\
     			\hline
     			IMU2 X&$-0.0947\pm0.0035$ & -0.0935 & \textbf{0.0012} \\
     			\hline
     			IMU2 Y&$0.0993\pm0.0015$ & 0.1010 & \textbf{0.0017} \\
     			\hline
     			IMU2 Z&$0.0011\pm0.0023$ & 0 & \textbf{0.0012}  \\
    			\hline
    			IMU3 Z&$-0.0007\pm0.0011$ & 0 & \textbf{0.0007}  \\
    			\hline
    			\hline
    			&Estimated$({}^\circ)$ & CAD Reference$({}^\circ)$ &Difference$({}^\circ)$\\
    			\hline
     			IMU2 Roll &$179.93\pm0.06$ &$180$ & \textbf{0.07} \\
     			\hline
     			IMU2 Pitch &$179.83\pm0.21$ &$180$ & \textbf{0.17} \\
     			\hline
     			IMU2 Yaw &$179.33\pm0.11$ &$180$ & \textbf{0.67} \\
    			\hline
    			IMU3 Yaw &$44.05\pm0.17$ &$45$ & \textbf{0.95} \\
    			\hline
		    \end{tabular}
		}
\end{table}

%%%%%%%%%%%%% implementation details
In practice, the cell's resolution is set at 0.5 m for indoor cases and 1.0 m for outdoor cases considering the scale of the environment. Additionally, in the first iteration, the cells with the plane-likeness coefficient $\mathcal{P}$ over 0.6 are accepted for data association. After the first batch optimization, motion distortion in LiDAR points will be compensated and a refined surfels map is available, we further improve the threshold of $\mathcal{P}$ to 0.7.
Except for the resolution of the cell and plane-likeness coefficient are different, other parameters remain unchanged in the experiment. Particularly, to render all quantities of the calibration observable and achieve more accurate calibration results, we ensure sufficient linear acceleration and rotational velocity over all datasets, and the knot spacing of the spline is set as 0.02 Sec to deal with the highly dynamic motion.

%%%%%%%%%%%%%%% vicon data
%%%%%%%%% TODO add a picture about compare the velocity
Comparing the estimated trajectories with the motion-capture system's reported trajectories, the average absolute trajectory error(ATE) ~\cite{grupp2017evo} over all ten sequences are 0.0183 m. Fig.\ref{fig:trajApe} shows two typical estimated trajectory aligned with the ground truth and also illustrates fitting results according to  Eq.~(\ref{Eq:ra}) and Eq.~(\ref{Eq:rw}). The estimated B-Spline trajectory fits well with the ground-truth trajectory and the acceleration measurements and the angular velocity measurements, which indicate the high representation capability of the B-Spline.

We also evaluate the repeatability of the proposed method on the calibration sequences collected from two typical scenarios. Table~\ref{table:mean-std-results}  shows the statistics over the transformation from IMU to LiDAR after eight iterations; the extrinsic rotations have been converted to Euler angle.
It can be seen that the final differences are within a few millimeters and milliradians in the indoor environment.
Due to the fact that the outdoor environment is more complicated and most planes are uneven, the repeatability of the proposed method over extrinsic translation outdoors is not as good as indoors. In general, the proposed method has a good repeatability across different sequences, especially for the calibration of rotation between sensors.

It is also interesting to study the effect of the number of iterations on the calibration results. As shown in the Fig~\ref{fig:8iterration}, the improvement between the first and second iterations is very significant. We think it origins from the poor initial surfels map and unreliable point-to-surfel data correspondences. After motion compensation, the surfels map and data association are more reliable. After about four iterations, the results gradually converge to the final values and remain unchanged.

Fig.~\ref{fig:cad} shows the final refined calibration results, the transformation from the IMU to the LiDAR, and CAD sketch for three IMUs. Table.~\ref{table:cad-results} details the relative poses results, where we set IMU1 as the origin and calculate the relative rotation and translation from IMU2 (or IMU3) to IMU1. The relative poses calculated from the calibration and the CAD are compared. The difference in relative translation is less than a few millimeters, and that in the rotation angle is less than $1{}^\circ$.

%% file: sections/07_Conclusion.tex
\section{CONCLUSIONS and Future Work}
In this paper, we propose a novel targetless LiDAR-IMU calibration method termed LI-Calib, based on continuous-time batch estimation. The real-world experiments indicate that the proposed method is highly accurate and repeatable, especially in the human-made structured environment. 
A current limitation of LI-Calib is the dependence on the NDT registration based LiDAR odometry in the first iteration. If the initial LiDAR odometry is poor, LI-Calib fails to get enough reliable point-to-surfel correspondences, which may lead to an inaccurate calibration. Consequently, it might be worth investigating some different front-ends for LiDAR odometry in the future.  

\section{Acknowledgement}
This work is supported by  the National Key R$\bm{\&}$D program of China under Grant 2018YFB1305900 and the National Natural Science Foundation of China under Grant 61836015. We owe thanks to Yusu Pan for the help of visualizing calibration results.